# Medical Technologies and Challenges of Robot Assisted Minimally Invasive Intervention and Diagnostics

Nabil Simaan, Rashid M. Yasin, and Long Wang *†

July 6, 2018

**Abstract**

Emerging paradigms furthering the reach of medical technology deeper into human anatomy present unique modeling, control and sensing problems. This paper discusses a brief history of medical robotics leading to the current trend of minimally invasive intervention and diagnostics in confined spaces. Robotics for natural orifice and single port access surgery, capsule and magnetically actuated robotics and microrobotics are discussed with the aim of elucidating the state of the art. Works on modeling, sensing and control of mechanical architectures of robots for natural orifice and single port access surgery are discussed, followed by a presentation of works on magnetic actuation, sensing and localization for capsule robotics and microrobotics. Finally challenges and open problems in each one of these areas are presented.

## 1 Introduction: Surgical Robotics from Open to Minimally Invasive and Robot-Assisted Surgery

The last two centuries have witnessed a slow progression toward minimally invasive surgery (MIS) as a surgical paradigm. In every disruptive phase of its evolution, MIS depended on new elements of technological innovation. The invention of the cystoscope allowed endoscopic exploration through the use of Bozzini's cystoscope[1] in 1805 and Desormeaux's endoscope in 1853 [2]. These two innovations supported

---

*Department of Mechanical Engineering, Vanderbilt University, Nashville, TN, USA 37204; email:nabil.simaan@vanderbilt.edu

Kelling's first laprascopic procedure in 1901 [3]. For several decades, surgeons were limited to peering through endoscopes using their own eye until the invention of digital cameras leveraged the invention of the Hopkins Rod endoscope. These two tools combined have offered a better view of the surgical field while supporting critical ergonomic requirements allowing surgeons to focus on manipulating instruments while looking at a monitor screen instead of peering through the lens of an endoscope. The availability of video-camera feed also allowed surgeons and assistants to simultaneously monitor the surgical site and to collaborate on surgical tasks.

The availability of the Hopkins rod endoscope and video-camera helped usher laparoscopic surgery starting with Kurt Semm's first laparoscopic appendectomy [4]. This has been characterized by increased adoption of minimally invasive technique for many surgical procedures including challenging ones such as the first Whipple procedure (pancreaticoduodenectomy) [5]. The resulting progress towards reduction of invasiveness has benefited patients by reducing blood loss, scarring, wound site infection, hernia, pain, and duration of post-operative recovery. The patient benefits due to the adoption of the laparoscopic MIS technique presented surgeons with several challenges when compared to open surgery. These challenges include a steep learning curve owing to the inverse kinematic mapping of hand-to-tooltip motion due to incisional constraints[1], lack of tool tip dexterity, and loss of sensory information. The need for manipulating multiple tools through multiple ports and the above-listed challenges have motivated the introduction of robot-assisted multi-port surgery, which grew steadily starting in the mid 1990's and growing rapidly after the first release of the da-Vinci system by Intuitive Surgical in the early 2000's.

The drivers leading to the current paradigm of computer-aided and robot-assisted surgery have been presented in [6]. Briefly, the desire to offer patients the benefits of MIS while sparing surgeons the technical difficulties associated with manual laparoscopic surgery and the desire to improve surgical outcomes by improving the accuracy of surgical execution have been the key drivers for computer-aided and robot-assisted surgery. These two goals have resulted in two ways in which technology has been used to help surgeons: *manipulation augmentation* and *perception augmentation*.

Perception augmentation was introduced through medical imaging followed by computer-aided navigation in order to improve surgical plan execution, to avoid accidental trauma to vital organs and to ensure complete excision of tumors. These advancements were enabled by the availability of computers and the first commercial ultrasound linear arrays, which accelerated the use of medical imaging in the early 1970's allowing the use of Ultrasound and Computed Tomography (CT) scannig [7]. These imaging modalities presented surgeons with tools for diagnostics, surgical pre-planning and intra-operative image guidance.

The concept of robot-assisted *manipulation augmentation* was introduced as a means to overcome the technical difficulties arising from the use of manual laparoscopic tools. The use of robotics decreased surgeons' learning curve who no longer had to contend with the reverse manipulation mapping of manual laparoscopy [8, 9]. Robot-assistance has leveled the field by reducing surgeons' physiological requirements (e.g. manual precision, steadiness and physiological tremor). Robots provided increased distal dexterity, allowed the manipulation of multiple arms, improved precision and steadiness and allowed multiple surgeons to manipulate multiple instruments and to collaborate on complex minimally invasive procedures. Advances in 3D stereo visualization and the development of dexterous distal wrists such as the Endowrist® (**Figure 1**) have helped reduce the cognitive and physiological burden associated with manual instrument laparoscopy. The availability of dexterous robotic wrists have enabled complex tissue manipulation and suturing that are very hard to achieve using manual laparoscopic tools. As a result, large swaths of surgical domains have seen wide adoption of robot-assisted MIS. For example, since 2003, more than 1.75 million robotic procedures have been performed in the United States according to a 2013 annual report by Intuitive Surgical[10]. In 2016 alone there were more than 750,000 procedures carried out worldwide by *da-Vinci* systems[11] with Gynecology and Urology being the surgical disciplines with the widest adoption (approximately 246,000 and 109,000 procedures carried out in the U.S. in Gynecology and Urology, respectively).

[1] Incisional constraints allow only 4 degrees-of-freedom (insertion along and rotation about the tool axis combined with two tilting motions about two perpendicular axes belonging to the local tangent plane of the skin at the incision point)

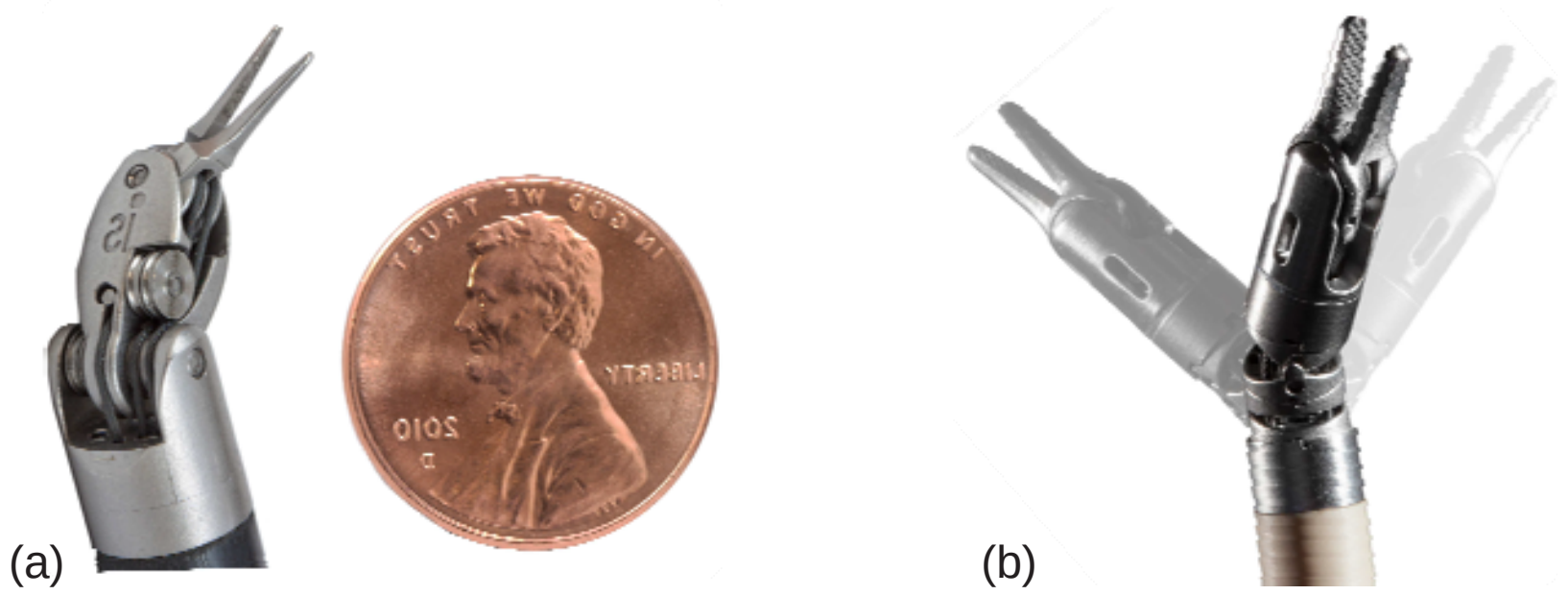

Figure 1: Dexterous wrists (Endowrist® by Intuitive Surgical): (a) 8mm wrist, (b) 5mm wrist

Surgical robotics has made significant strides in changing patient care. However, there are still many areas where adoption of robotics is limited due to limited patient outcomes justifying the increased surgical difficulty, cost and risk due to the MIS approach compared to open surgery [12, 13]. Furthermore, minimally invasive systems still heavily rely on using rigid instruments with dexterous distal wrists, which have been noted for their limitations when considering new surgical paradigms for accessing internal anatomy without skin incisions (e.g. natural orifice surgery) or by using a single incision [14, 15, 16, 17, 18, 19]. The continued demand to allow deep access into the anatomy has motivated researchers in the past decade to explore three key areas: dexterous snake-like robots for surgery, magnetically actuated devices and microrobots. This paper provides an overview of progress made in these areas while first addressing the key technical challenges presented by the newly emerging surgical paradigms of natural orifice surgery and single port access surgery.

# 2 Challenges of Robot-Assisted Surgery & New Frontiers of Surgery with Confined Access and Perception

Though current robotic systems are able to address the manipulation requirements for a very large set of surgical applications, the adoption of robot-assisted MIS over open surgery or manual laparoscopic surgery has not gained wide acceptance across all surgical disciplines. Beyond the socioeconomic reasons and difficulties in carrying out cost-benefit analysis in light of post-operative outcomes, there are key technical hurdles that explain this low rate of adoption. We will limit the discussion here to the technical hurdles.

Minimally invasive surgery can be categorized into MIS in shallow and large spaces and MIS in deep surgical fields and confined spaces. For example multi-port trans-abdominal MIS of the abdominal cavity presents a large workspace for instruments to maneuver and is considered MIS in a shallow and large space. On the other hand, MIS of the upper airways is MIS in confined space. The two categories present substantially different technical challenges in terms of demands for manipulation and sensory augmentation. To understand the technical hurdles to wider adoption of robot-assisted surgery across all surgical disciplines, we will first illustrate the limitations of MIS in shallow and large workspaces, thereby explaining the limitations of current commercial surgical systems for multi-port MIS. We will then discuss the additional challenges associated with MIS in deep and narrow workspaces within the context of the new emerging surgical paradigms for natural orifice and single port access surgery.

## 2.1 Challenges of Robot-assisted MIS in Shallow and Wide Spaces

Removing the surgical tools from the surgeon's hands allowed surgeons to exceed physiological limitations by harnessing the advantages of multi-port robot-assisted MIS, which have been outlined above. However, taking the tools from the surgeon's hands results in limited sensory perception and situational awareness. When operating tools using current commercial robotic systems surgeons have to operate with obfuscated sensory feedback. They cannot feel tool interaction with the anatomy since current systems do not offer force feedback. Also, they cannot see a full view of the surgical field allowing for fast and easy formation of understanding of the surgical scene since surgical site visualization is achieved through the use of narrow baseline stereo vision cameras. These two technological gaps are a pre-requisite to enabling the full benefits of dexterous, safe and high-precision manipulation free of surgical errors or accidental trauma to the anatomy.

**Figure 2** shows the da-Vinci Xi system for multi-port MIS, a typical abdominal MIS setup with several trocars penetrating the abdomen to provide tool access and a listing of the key challenges that multi-port MIS presents to surgeons. Existing systems can handle most of these challenges through multi-arm designs that also incorporate dexterous wrists. However, the challenges of limited visualization and the sensory deficiency associated with use of current robotic systems can impact the surgeon's ability to carry out surgical tasks as easily as in open surgery. These challenges present situational awareness barriers that limit the surgeon's ability to interpret the surgical scene, to associate the scene with preoperative imaging information and to safely complete surgical interventions.

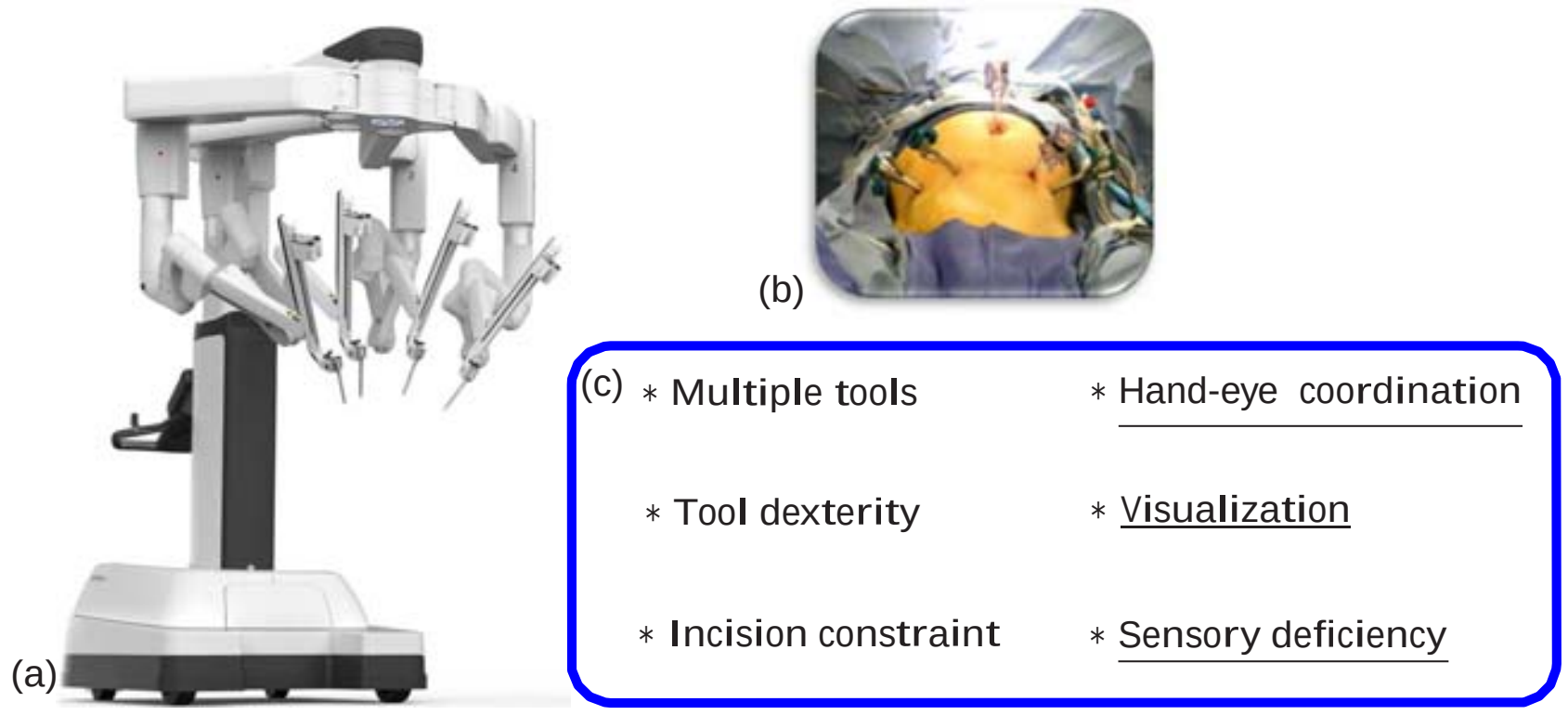

Figure 2: (a) The da-Vinci Xi® multi-port system by Intuitive Surgical, (b) a typical multi-port setup for abdominal MIS and the challenges of multi-port MIS in shallow and wide spaces, (c) challenges limiting surgical situational awareness are underlined.

## 2.2 Challenges of Robot-Assisted MIS in Confined and Narrow Spaces

In an effort to further reduce post-operative pain, risk of hernia, wound site infection, scarring and formation of adhesions, surgeons have proposed the paradigms of natural orifice surgery and single port access (SPA) surgery. The vast majority of natural orifice procedures use an *endoluminal* approach during which the surgical site is reached by traversing and operating within an anatomical lumen. Examples of this approach include trans-urethral bladder cancer resection, trans-oral upper airway surgery, trans-esophgegeal gastric surgery and trans-anal endoscopic microsurgery. In other scenarios *transluminal* natural orifice endoscopic surgery (NOTES) is used whereby surgical tools traverse anatomical passageway and break through the access lumen to reach a surgical site contained in another anatomical space. Examples of NOTES include trans-vaginal abdominal surgery, transsphenoidal petuitary gland surgery and trans-gastric abdominal surgery. During SPA a single access port (typically placed at the umbilicus) is used to provide access for all surgical arms into the abdominal cavity.

Both SPA and natural orifice surgery present unique challenges in addition to the challenges of multi-port MIS. These challenges are briefly summarized in **Figure 3**. The figure shows the da-Vinci single site system developed for SPA along with an illustration of natural orifice intraluminal endoscopic surgery where a long surgical tool traverses a long and narrow anatomical passageway and an example of a SPA robot operating within a bodily cavity. These challenges stem from the need to operate multiple tools at a confined space and the need to have multiple tool shafts converging through a narrow access path. For example, the da-vinci single site system in **Figure 3-(a)** tries to address the latter need by using curved instruments at the expense of distal dexterity loss. Unlike in multi-port and single port transabdominal MIS where the rigid robotic surgical arms contact the anatomy only at their gripper and at a predetermined fulcrum point through the trocar, robots for natural orifice surgery have to traverse complex anatomical passageways and they may contact the anatomy at multiple points/regions along their length as illustrated in **Figure 3-(b)**. Finally, in NOTES procedures (e.g. trans-gastric abdominal surgery) there is the significant challenge of obtaining wound closure within the gastric wall after completing the procedure.

In addition to the challenges above, MIS in confined spaces presents additional challenges further limiting the surgeon's situational awareness compared to MIS in shallow and wide spaces. These challenges include limited perception of the surgical scene and the robot and complex tele-manipulation mappings. **Figure 3-(c)** illustrates the perception barrier imposed by the use of endoscopes in confined spaces. Compared to multi-port MIS, the visible portion of the anatomy and the robot is more limited and possible collisions between the robot and the anatomy are masked. Also, compared to MIS in shallow and wide spaces where generally there is a correspondence between the motion range and shape of wristed surgical tools and the surgeon's hand (e.g. as is the case for the da-Vinci Endowrist in **Figure 1**), in NOTES the robots must have many degrees of freedom and highly dexterous arms and this correspondence become significantly more complex to learn.

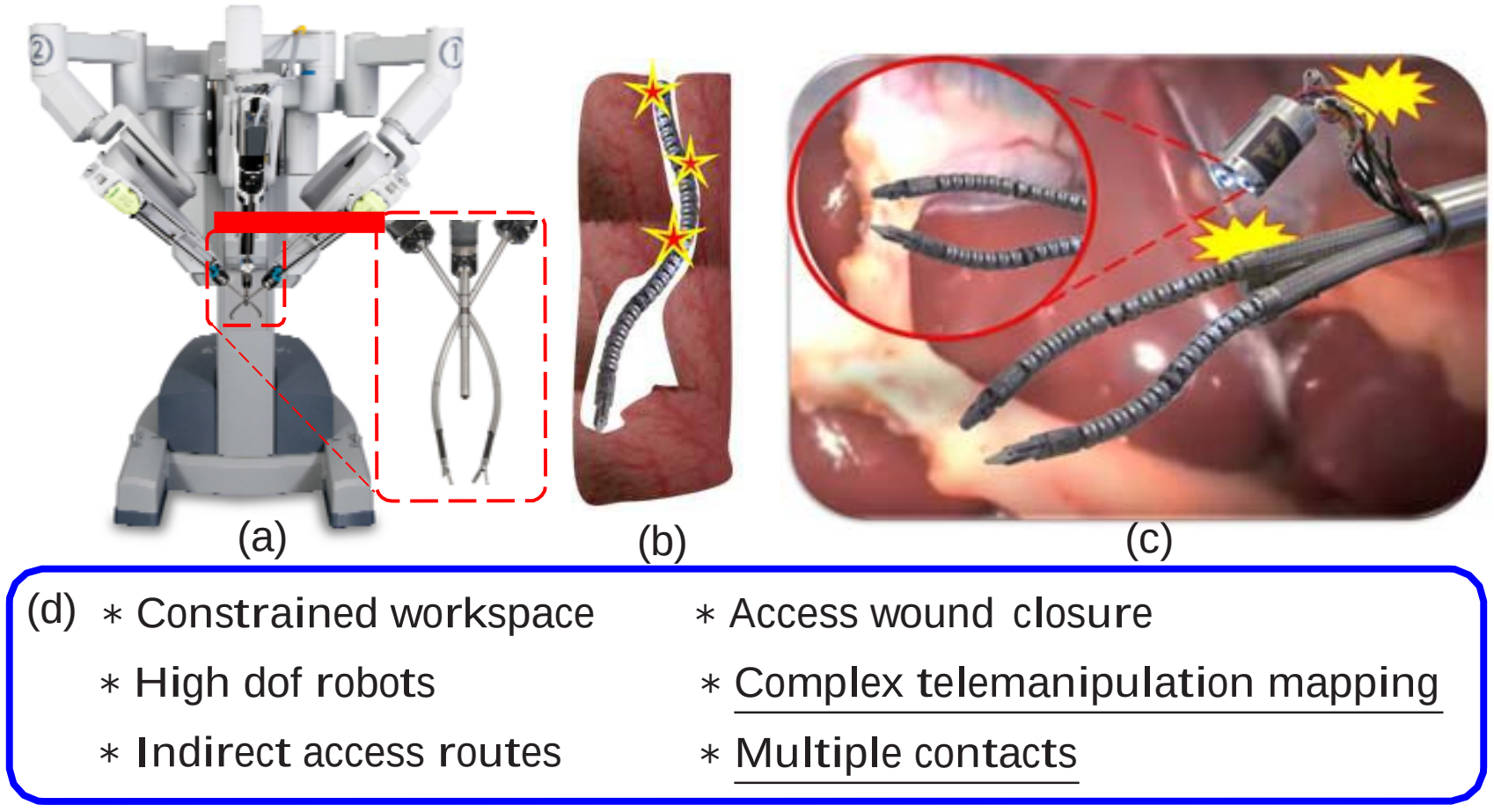

Figure 3: (a) The da-Vinci Xi® multi-port system by Intuitive Surgical, (b) Multiple contacts along the body of surgical arms during NOTES, (c) Example of limited field of view during SPA surgery, (d) Challenges that MIS in confined spaces presents in addition to the challenges shown in Figure 2. Challenges limiting situational awareness are underlined.

# 3 Technologies for Diagnostics and Intervention in Confined Spaces

There are several approaches to minimizing or eliminating the number of access incisions for surgical intervention. In the following, we review surgical systems for NOTES and single port access surgery, systems that use magnetic actuation, capsule robotics and micro-robotics. These reviews aim to outline the progress made and to help elucidate the limitations and challenges in each area.

## 3.1 Surgical Systems for NOTES and Single Port Access Surgery

The robotic systems used to address the challenges of NOTES and SPA present complex mechanical architectures due to the stringent operational constraints and high number of actuated joints used to provide the necessary kinematic dexterity in confined spaces. Detailed reviews of these systems have been included in [20, 21, 22, 23, 24]. In this section, we briefly present an overview of the mechanical architectures used to the extent needed to outline their associated modeling and control challenges. A more detailed review of these robotic architectures can be found in [25].

High dexterity snake-like robot architectures are typical in systems designed for NOTES and SPA. These robots come in many designs that can be categorized as a) articulated with embedded actuation b) linkage-based designs, c) wire-actuated designs and d) designs using continuum robots. **Figure 4** shows examples of such systems. **Figures 4-(a-b)** show two examples of a NOTES robot [26] and the SPRINT robot for SPA surgery [27]. These robots embed the electromagnetic actuators within their dexterous arms. **Figures 4-(c-d)** show two systems for SPA surgery that predominately relay on linkages for delivering actuation to their end effectors [28, 29]. **Figures 4-(e-f )** show the da-Vinci single port surgical system and the SAIT-KAIST single port surgery system [30]. Both of these systems use articulated serial structures relying on wire actuation to control their end effectors. **Figures 4-(g-h)** show the insertable robotic effectors platform (IREP) for SPA surgery [31] and another system for transuretrhral laser-enucleation of the prostate [32]. The IREP uses a multi-backbone continuum architecture in which push-pull actuation is used on a multitude of super-elastic NiTi backbones in order to affect a desired bending shape of each segment. The robot in **Figure 4-(h)** uses a concentric tube design in which a stack of concentric superelastic NiTi are rotated and translated with respect to each other in order to achieve a new equilibrium shape.

A common feature for all these dexterous systems[2] is the use of distal dexterity wrists and multiple arms capable of coordinated motion at the same site (i.e. triangulation of arms). The key function of the distal wrists is to enable dual-arm passing of circular needles and knot tying in confined spaces. One mode of operation in which the last active joint of the wrist includes rotation about the longitudinal axis of the gripper is of particular importance for NOTES and MIS in confined spaces. **Figure**

[2]With the exception of systems using concentric tube continuum robots (e.g. [33] or access over-tube platforms for NOTES such as the HARP robot [34]

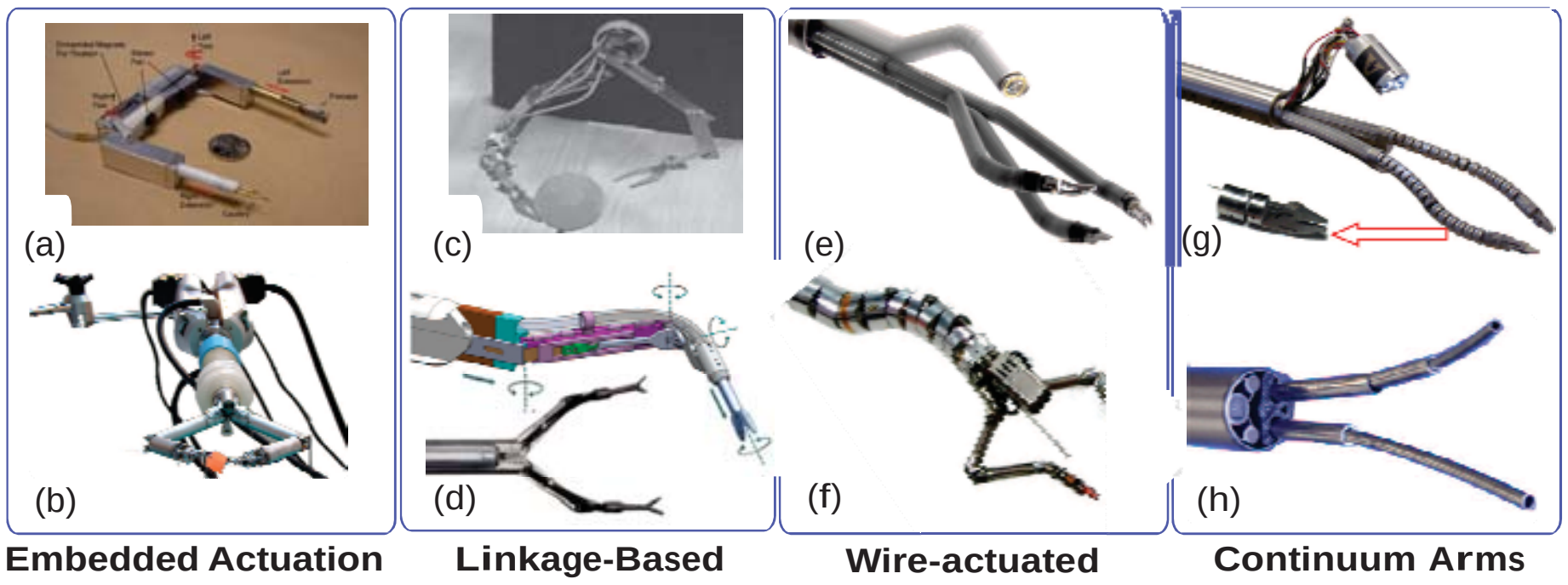

Figure 4: Various actuation methods for robots for NOTES/SPA: (a) An insertable NOTES robot with embedded actuation [26], (b) The SPRINT system for SPA [27], (c) Six-bar insertable system for SPA [28], (d) SPA system using torque tubes and linkages [29], (e) da-Vinci single port system using wire actuation, (f) The Samsung Advanced Institute of Technology wire-actuated SPA system [30], (g) the IREP robot with multi-backbone continuum arms, (h) A handheld robot using concentric tube continuum arms for transurethral laser prostate surgery [33].

**5-(a)** shows how suturing in confined spaces can be accomplished by passing circular needles through the rotation of a rigid needle holder. This mode of operation inspired two methods for achieving roll about the longitudinal axis of the gripper by using the body of the robot to transmit rotation about its backbone (see **Figure 5-(b)**) or by designing a roll wrist at the distal end of the robot (**Figure 5-(c)**). **Figure 5-(d)** shows a sequence of images depicting knot tying and roll about the backbone of a multi-backbone continuum robot [35, 36]. An approximation of this mode of operation has also been demonstrated in the recent da-Vinci single port surgical system. **Figure 5-(e)** shows the IREP robot using its roll wrist to carry out knot tying. This same design concept was presented also in the Titan SPORT™ system by Titan Medical and in the Samsung Advanced Institute of Technology (SAIT) and Korea Institute of Science and Technology (KIST) single port access system [37].
The above-listed designs come close to meeting the dexterity requirements of NOTES and SPA surgery. However, there are several associated modeling, sensing and control challenges that arise from the use of design architectures such as in **Figure 4**. These challenges are outlines in section 4.1.

## 3.2 Closed-Loop Control and Calibration of Wire-Actuated and Continuum Robots

One key characteristic of NOTES and SPA systems is the use of wire actuation and continuum arms/linkages with significant deflections and frictional and motion losses. Due to these losses, the motion control accuracy of these surgical systems is a challenge. Many works have focused on improving the position control accuracy of such robots during surgical procedures. The majority of works introduced extrinsic measurements into the control loop to close the error in an online fashion. Jayender et al. demonstrated position control of active catheter using real-time image tracking and using an electro-magnetic sensor [38, 39]. Penning et al. proposed and evaluated a closed-loop position control framework for robotic catheters based on inverse kinematics [40, 41]. Bajo and Simaan proposed and demonstrated mixed feedback approach using joint and configuration space measurements for improving motion tracking of multi backbone continuum robots [42].

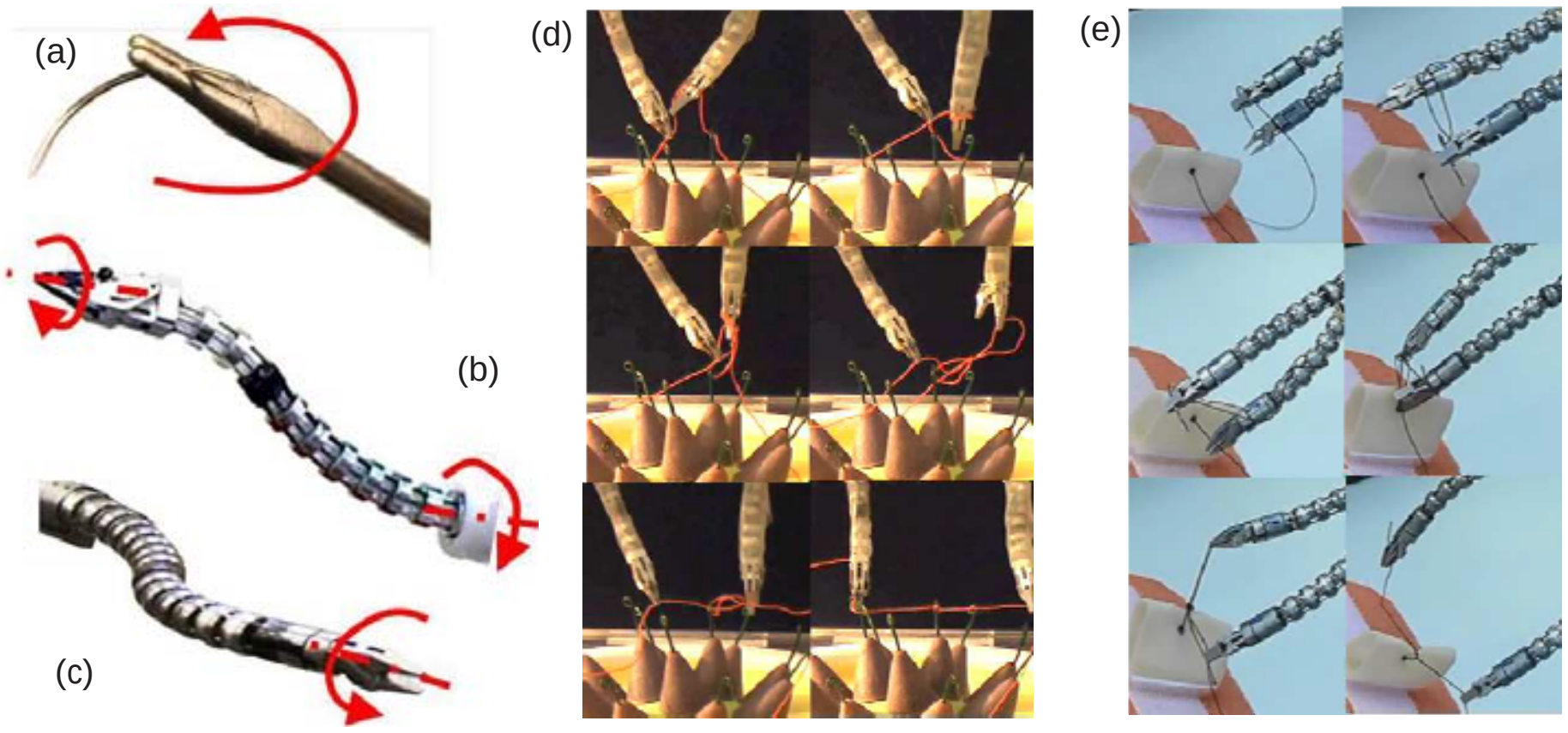

Figure 5: Suturing in confined spaces through rotation about the longitudinal axis of the gripper: (a) a needle holder, (b) a continuum robot transmitting rotation about its backbone, (c) a continuum robot with a dedicated roll wrist, (d) knot tying using rotation about the backbone [35], (e) knot tying and needle passing using a dedicated roll wrist [31]

In addition to approaches aiming at using online measurement for improved motion tracking, there have been approaches aiming at improving the accuracy of the kinematic models of continuum robots via calibration oe experimental characterization. For example, the shape and torsion of concentric tubes were calibrated using constant curvature assumption in [43]. Coupling effects in wire-actuated two-segment catheters were calibrated using vision in [44]. Kinematics of a flexure-based single backbone continuum robot was characterized using image analysis and constrained optimization in [45]. In [46] a parameterized modal approach was used to formulate the bending shape deviation and coupled twist and to investigate the error proropagation for multi-backbone continuum robots More recently, a generalized Jacobian for concentric tubes toward on-line parameter estimation using an extended Kalman filter was presented in [47].

## 3.3 Friction and Extension of Tendon/Backbone Actuation Lines

Robots designed for natural orifice and minimally invasive surgery are mostly wire actuated (e.g. [48, 34]), with the exception of recent growth on continuum robots (e.g. [49, 50, 51]). These robots locate their actuators remotely from their end effectors in order to facilitate miniaturization. For example, **Figure 6-(a)** shows Bowden cable actuation. **Figure 6-(b)** shows actuation via continuum backbones with an example of the IREP system as a particular case study for motion and frictional losses. These losses due to compliance and motion losses and friction in the actuation lines adversely affect end effector motion accuracy [35, 52], force sensing capabilities [53] and the control stability [54].
To model force transmission losses due to friction, various approaches have been proposed and applied. A simple transmission model assuming Coulomb friction was presented in [54] and a dynamic friction model was used in [55] to incorporate viscoelastic properties of tendons. Menciassi's group [56, 57] presented a friction loss model based on the capstan friction model with discrete approximation of the shape of the Bowden cable pair using constant radii of curvature to approximate varying curvature. A continuous time-domain model of tendon-sheath transmission was to describe a conduit of non-uniform curvature profile using similar discretized approach in [58]. Conduit-induced friction was incorporated into the kinematic model of a multisection tendon-driven robot in [59]. More recently, [60] proposed a modeling framework that allows estimation of internal friction parameters in varying environments and in varying sheath shapes.

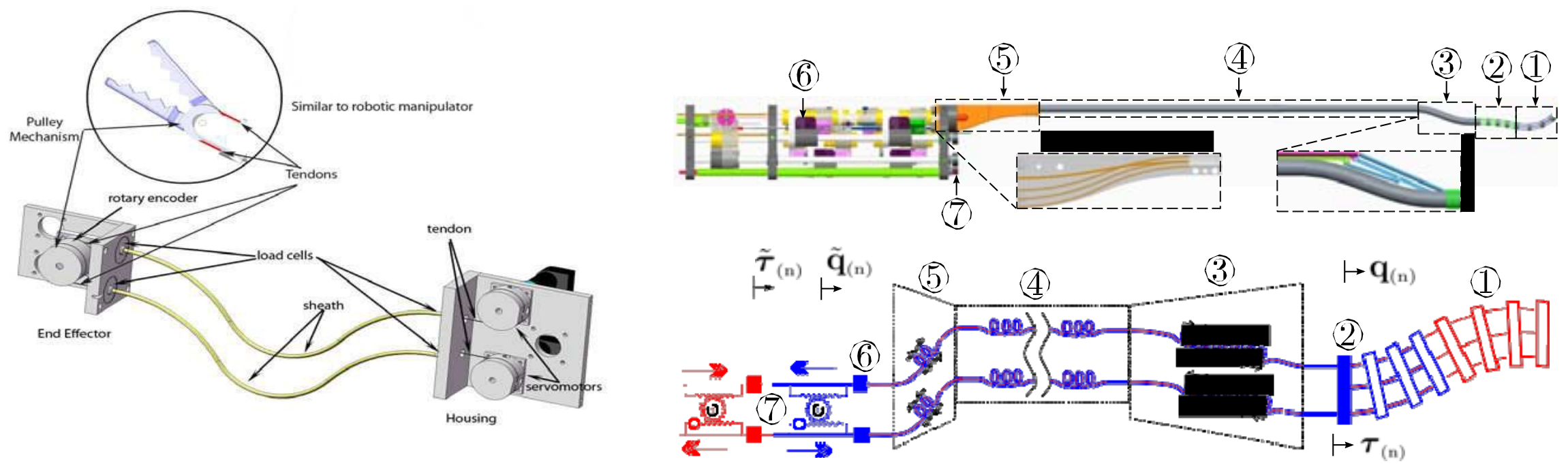

(a) courtesy of Chiang et. al 2009

(b) courtesy of Roy et. al 2017

Figure 6: Examples of modeling works on friction and extension of tendon/backbone based actuation lines: (a) concept sketch of the framework proposed in [56] for sheath modeling in tendon based actuation systems; (b) an illustration of the actuation system of a multi-backbone surgical continuum robot and its modeling elements as in [60]

To compensate actuation line extension, several modeling and control methods have been proposed. Xu and Simaan [61] proposed recursive linear estimation using measured actuator joint position and measured end effector pose to capture backlash and correction factors for elastic properties. Simaan et al. [35] extended this work by characterization of motion transmission losses and coupling using Fourier series and feed-forward actuation compensation using the statics model of a multi-segment multi-backbone continuum robot. Agrawal et al. [52] proposed and tested the use of the tangent hyperbolic smooth inverse for backlash feed-forward compensation in wire actuated robots. Actuation compensation of catheters was proposed in [62], using Coulomb friction model and a characterization of the backlash as a dead-zone function. Roy et al. [60] extended these works by presenting an approach for modeling motion and frictional losses in multi-backbone robots and presenting an approach for identifying these parameters in a case study for the IREP system.

## 3.4 Sensing of the Interaction Between Robot and Environment

Failure to control surgical interaction forces could result in tissue damage, limited blood flow to sutured organs [63, 64], or leak of bodily fluids around loosely sutured internal organs [65]. Hence, many studies [63, 66, 67, 68, 69, 70], have investigated force feedback in telemanipulation and have shown that it reinstates the surgeon's ability to perceive and control tool interactions. Technologies used for in-vivo force sensing include instrumented laparoscopic tools (e.g. [71, 72, 73, 74, 75]).
The force sensing elements used in MIS were reviewed in [76], including strain gauges [73, 77, 78], load cells [71], LVDT sensors [79], fiber Bragg grating and optical micrometry [80, 81]. While instrumented surgical tools with tooltip force sensors are conceptually attractive, they present significant challenges in terms mechanical ruggedness, sterilization, cost, and MRI compatibility. As an alternative to expensive tooltip force sensors, other researchers have tried to estimate tooltip forces by monitoring actuator effort using motor current monitoring [82], strain gauges [77, 83, 84], pneumatic pressure monitoring [72], dithering while monitoring actuation forces [85, 56], and using the joint actuator forces [86, 53].

## 3.5 Magnetic Actuation and Fixation for Surgery

One area of active research in further reducing invasiveness in minimally invasive surgical systems is the use of magnetic technologies for actuation and fixation of instruments to the abdominal wall. These methods propose to reduce invasiveness of laparoscopic or NOTES procedures by passing tools through a transluminal incision or a single abdominal incision and then affixing these tools to the patient's tissue with an external magnet. Most works focus on tool fixation across the abdominal wall, spanning a range of different form factors. Early systems showed the use of a single tool for retraction or endoscope orientation, but more recent works have combined a variety of tools to allow multi-functional surgery that is controlled and mounted to the body magnetically. A recent review focused completely on magnetic surgical instruments for abdominal surgery [87].

### 3.5.1 Types of Magnetic Systems

Early works showed the ability of an external permanent magnet to control the spatial position of a single tool on the inside of a patient by coupling an external magnet with a magnet at the base of the tool. This can be used to pull traction on a clip attached to anatomy [90], to position a camera [91], or to retract tissue [92]. The gross motion achievable by manipulating an external magnet is not suitable for very fine manipulation of a device and has mostly been used as an anchoring method for tools with additional internal actuation modalities.
There is an array of different tool types that can operate anchored to the abdomen with a magnet. Systems include: a 3 DOF pneumatically actuated cauterizer [93]; a magnetically anchored pan tilt and translation camera system [91, 94]; various retractors [95], and a number of multi-degree of freedom arms [88, 26, 96]. Some of these works have incorporated the use of reconfigurable frames that insert with the tool and keep the tools mounted together in order to localize the arms with respect to one another and to provide better dexterity. All of these systems are intended to be inserted into the body through a single incision. Some [93, 94, 91] were intended for single incision laparoscopic surgery, whereas others [97, 95, 26, 96] were made to be compatible with a transgastric incision after transoral insertion of the device.

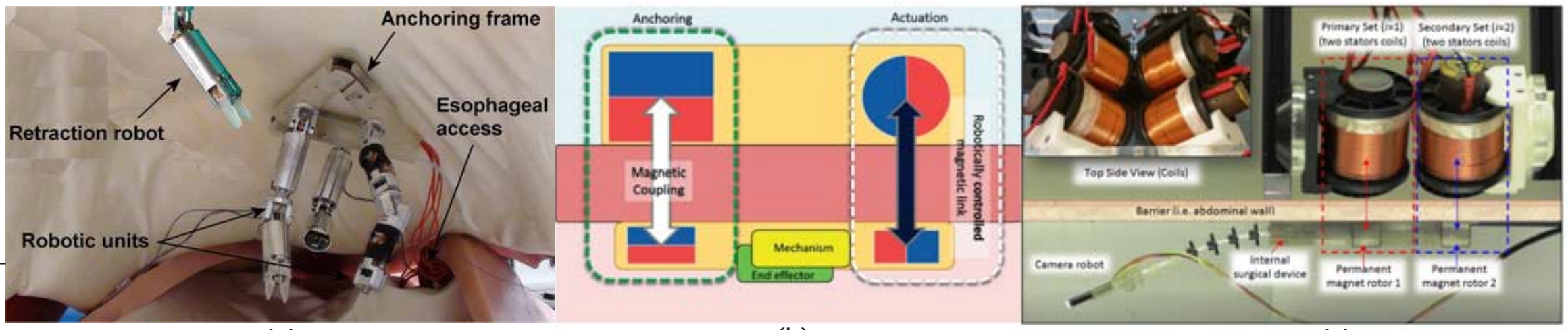

(a) (b) (c)

Figure 7: Magnetic actuation across the abdominal wall. (a) Magnetic anchoring frame with internal electric actuation.(b) The left set of magnets are used to anchor and perform gross movement of an instrument, with the right using spinning circular magnets to generate rotary motion. (c) Similar to (b), external wire coils generate a time-varying magnetic field to spin a magnet inside the abdomen. Images courtesy [88, 87, 89]

### 3.5.2 Magnetic Actuation Methods

The combination of magnetic anchoring with internal DC motors is promising, but limited by the power of devices available in such small packages and the need to pass current-carrying wires through the access port. This has motivated further work in magnetic actuation to be used in conjunction with magnetic anchoring for a wireless NOTES system not relying on internal DC motors.

Magnetic actuation can be achieved in a variety of ways. The simplest is to use an altered version of the anchoring method: by moving one anchor with respect to another, rotary motion can be achieved via a linkage [93]. Another method is the use of a shielding material between the external and internal magnet to allow control over the applied force between the magnets [98]. To achieve rotary motion, rotating magnets can be used to couple internal and external permanent magnets to transmit torque from an external circular magnet to an internal magnet that transmits torque in a similar fashion to a spur gear [99], as seen in **Figure 7-(a)**. An example closed-loop actuation scheme using coupled internal and external permanent magnets showed transmission of 13.5 mNm of torque in closed loop actuation [100]. Actuation of rotary joints inside the body can also be achieved electromagnetically in a configuration similar to an electrically commutated motor: wire coils outside the body can generate magnetic fields and act as stators located outside the body to turn a permanent magnet acting as a rotor inside the body, as seen in **Figure 7-(b)**[101].

## 3.6 Robotic Capsules for Diagnostics and Intervention

Deep access into the anatomy has taken off for natural orifice surgery most prominently in inspection of the gastrointestinal (GI) tract. As an alternative to traditional GI endoscopy, a camera pill was first developed in 1981, larger experimental trials carried out in the mid 90's, and swallowable systems were developed in the early 2000's by commercial companies. While adoption is not as complete as traditional endoscopy, by 2009 more than 750,000 patients had undergone wireless capsule endoscopy [102]. On the market now, there are a number of different commercial systems for capsule endoscopy across the globe, with well over one million capsule endoscopy patients.[103]
The basic operation of these commercial systems is similar: a swallowable capsule relies on wide-angle cameras to see as much of the anatomy as possible while passing through the GI tract and recording or transmitting video. In the stomach, magnets or swimming techniques can be used for orienting capsules in a fluid-filled hollow organ, but for the lower GI tract, commercially available capsules currently rely on peristalsis for their movement. This motivates an active area of research for locomotion and localization throughout the entire GI tract. Other research focuses on the ability to combine the visualization abilities of currently sold capsule systems with therapeutics and better localization abilities. A 2017 review highlights many of the important developments in the field up until the present [104].

### 3.6.1 Capsule Locomotion

For capsule-based systems, locomotion methods are largely split between mechanical and magnetic methods. A full review specifically targeting locomotion systems was performed in 2015 by Liu et al[105]. One particularly novel system seen in **Figure 8-(a)** focused on using electrodes to stimulate the intestinal muscles to speed up peristaltic motion, tested in a single human subject in 2005, but a later trial found decreased heart rates during a test with a volunteer. [106]
In the stomach, flexible endoscopy is the preferred diagnostic method as it allows active actuation to view the entire volume of the stomach - passive capsules are not able to view enough of the anatomy. To investigate potentially less invasive methods, swimming-based and magnetic capsule actuation methods have been developed using liquid stomach distension, see **Figure 8-(d)** [97]. Early trials reported the use of a handheld magnet to orient a capsule with another magnet inside a water-distended stomach, but these methods are largely nonintuitive to use, leading to the use of robotic methods to reduce fatigue and improve usability/precision [104]. An alternative to the use of moving permanent magnets is electromagnetic coil systems. For example, Siemens and Olympus have developed the Magnetically Guided Capsule Endoscopy technique that uses 12 coils to guide a swimming capsule in the stomach. In the intestine, mechanically actuated systems have included worm-type, legged, wheeled, or crawling systems, with some examples in **Figures 8-(b-c)**. Actuation relies on shape-memory alloys, piezoelectric actuators, stepper-motors, DC motors, or pneumatics [107, 108]. However, power requirements of many of these systems requires external tethering for electric or pneumatic power, providing a large drawback.

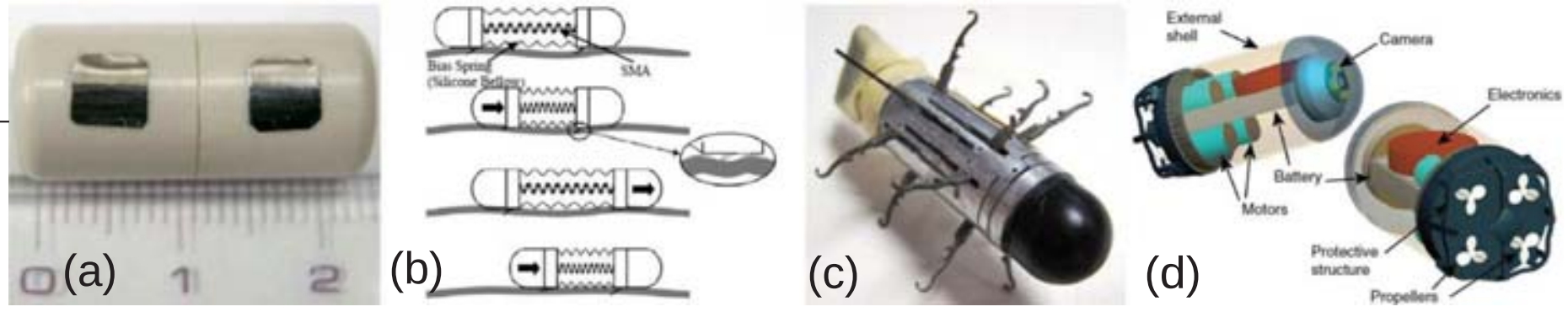

Figure 8: Wireless capsule actuation methods. a) Electrode-based locomotion for inducing increased peristaltic motion [106] b) Shape memory alloy "Earthworm" locomotion [107] c) Crawling capsule with hooked legs [108] d) Propeller-based swimming motion[97]. Image mosaic courtesy of [104]

The high power requirements as well as concerns with fragility, mechanical complexity, sterilizability, and possible organ trauma have resulted in more recent research moving away from internal actuation methods to magnetic actuation methods.
There are a variety of categories of magnetic actuation methods: single magnet, moving magnet, multiple magnet, electromagnet, fringe field, or magnetic resonance navigation. These have varying costs, field strength, accessible degrees of freedom, and complexity to implement, as summarized in [109]. Magnets can be used to drag/orient capsules [110] or to create internal motion for some sort of swimming motion. This includes flexible-tailed fish-like capsules or rotation-induced spiral actuators [104].
In the field of control of magnets more generally, telemanipulation of magnets using external electromagnets has been achieved with a linear quadratic gaussian controller, haptic feedback with virtual fixtures, and force sensing using hall effect sensors on a capsule [111, 112, 113].

### 3.6.2 Capsule Sensing

A variety of different sensing modalities have been proven useful for anatomical monitoring when integrated in capsules, summarized by Shamsudhin et al in their 2017 review [104]. In gastric environments, multisensing capsules have been developed to measure pH, pressure, and temperature to estimate gastric motility and transit times or esophageal pH. Capsule sensors have also shown the potential for acoustic heart and breathing rate monitoring in the GI tract in a porcine model through the use of a microphone. GI measurements of transluminal impedance, electrical permittivity, or bleeding have also been integrated.
Localization of capsule robots is accomplished in a variety of methods. For capsule endoscopy, video feeds and anatomical landmarks give a rough indication of anatomical location, but no specific telemetry information. High-technology, high-cost methods have been developed to find capsules in MRI, CT, or PET scans, which are less likely to adopt throughout endoscopy of the entire GI tract due to scanner costs and radiation doses.
To find more cost and time-efficient methods that can also be used with external actuation sources, there area a number of works on electromagnetic tracking of capsule robots relying on either external or capsular sensors [114, 115, 116, 117, 118]. New research has capitalized on results of these works in order to enable simultaneous magnetic actuation and localization - many of the previously investigated methods for localization would not be compatible with magnetic actuation and the presence of strong magnets would similarly be infeasible or have negative effects on other localization methods.
Magnetic field sensors can be used to localize capsules with point dipole models while they are under the control of an external magnet [119, 120]. However, due to nonlinear effects of the magnetic field, these works have difficulty when the external magnet is close to the capsule which is desired during capsule manipulation. Methods relying on measurements after aligning the capsule with the external magnet can achieve low error when the magnet is close to the capsule [120], but require the ability to always realign the capsule in the same orientation of the external magnet which may not be a valid approach in all anatomical situations.
Revolving one or multiple external magnets around a capsule with magnetic sensors [119] has allowed for simultaneous localization and control of a capsule in 5 degrees of freedom with average errors of 11 mm and $11^\circ$. Others have fused inertial and magnetic field sensing with estimation errors around 5mm and $19^\circ$ with interactive refresh rates around 50 Hz [121]. More recent work [122] was able to achieve greater than 100 Hz measurements with an average error in closed loop control of a capsule below 7 mm and $5^\circ$.

## 3.7 Medical Microrobotics

Working below the scale of the relatively large capsules in the GI tract to smaller microrobots would allow for minimally invasive treatment of the circulatory system, the urinary tract, the eye or the nervous system. However, the small scale of such devices leads to a new set of challenges to overcome. As shown in **Figure 9**, therapy options for such robots includes targeted therapy (brachytherapy or drug therapy), material removal (ablation or biopsy), telemetry (sensing or marking), and the introduction of controllable structures (scaffolds, stents, implants). Microrobotic sensors could be used for continuous sensing and health monitoring or blood pressure, muscle/neurological activity, etc.

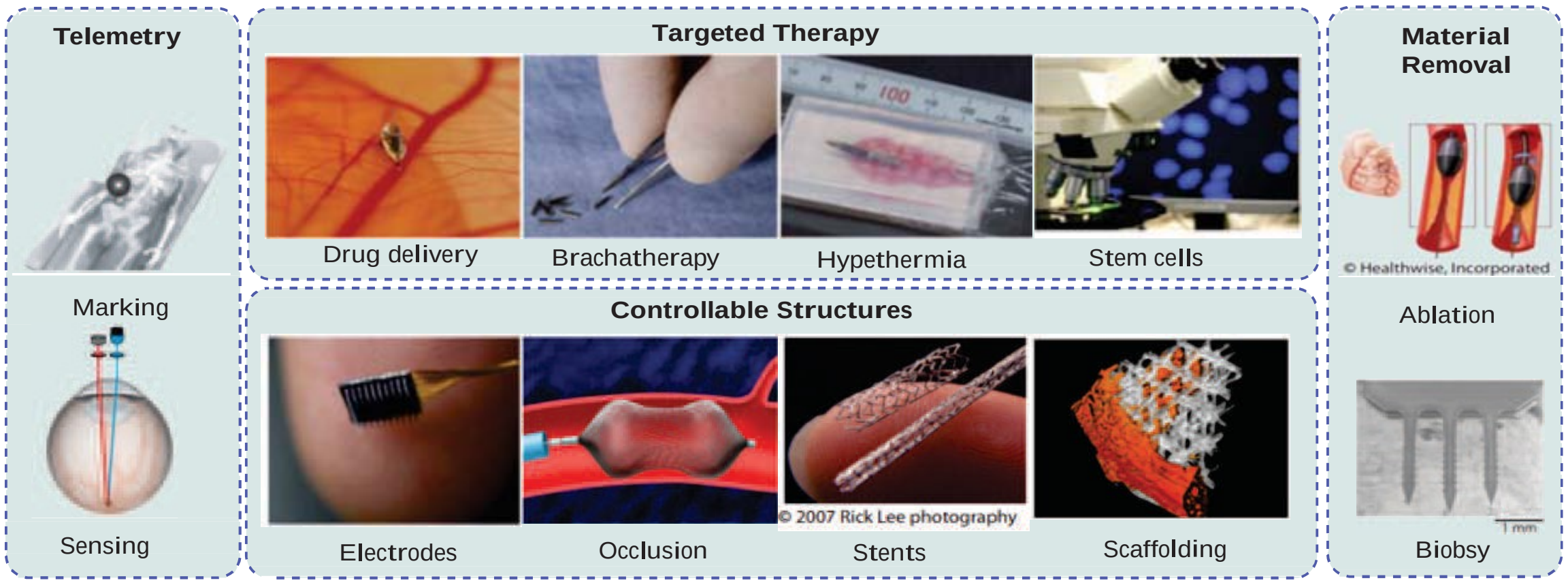

Figure 9: Various applications of micro robotics. Edited with permission based on [123]

There are a variety of potential biomedical applications for these microrobots. A number of review papers have helped summarize some of the main contributions to this field [124, 123, 125]: Targeted therapy for drug delivery has been shown possible with mechanical delivery methods, light, electric heating, or magnetic drug ejection. Small micro-biopsy devices have been developed to take tissue samples. Researchers have shown the possibility of using microrobots as scaffolds for tissue regeneration or in vitro tissue growth. They have also been shown to be able to target and transport or cut individual cells.

Some particular uses of microrobots include the use of millimeter-scale origami robots with magnetic locomotion for targeted drug release or ingested battery removal [126] and clogged vessel clearing with a helical microrobot that achieved mechanical clot grinding in an in vitro model [127].

Locomotion can be achieved through helical flagella driven magnetically or travelling-wave flagella driven with piezoelectric motors or external magnets. Propulsion is difficult due to the low Reynolds number of interactions at small scales which makes some locomotion/swimming techniques infeasible, leading to new developments of miniature piezoelectric motors with flagella or swimming devices powered by external magnets [125]. Similar to the control of capsules with magnetic fields, electromagnetic coils can be used to control magnetic field shape and strength for anchoring, moving, or orienting internal magnets or ferromagnetic structures inside hollow cavities: for example creating 5 DOF motion of swimming robots inside the eye [128]. External magnets or MRI machines can also be used to directly pull microrobots.

Battery technology does not at this time apply well to the micro scale, so is mainly applicable for robots on the larger end of the spectrum. To counter this deficiency, external power sources or energy harvesting techniques have been devised to power these robots. MEMS power generators have been developed that convert mechanical or thermal energy. Chemical harvesting has been tested for biofuel cells. Microorganisms can be co-opted to serve as propulsive agents. Power can also be transmitted through induction or provided externally by magnets for locomotion, or via RF power transmission [125, 123].

Localization of these systems can be achieved visually for in vitro tests or those systems used in the eye or similar other externally visible organs, but in more internal organs other methods are still under exploration. Recent reviews [129, 130] have elucidated some of the recent progress toward better localization of medical microrobotic systems. For truly micron-scale systems, ultrasound lacks the fine resolution to accurately find these systems. Positron Emission Tomography (PET) has the ability to find small structures, but comes with a high cost, more invasiveness due to the presence of radioactive markers, and high cost; computed Tomography has many of the same problems. Magnetic Resonance Imaging (MRI) has the required resolution and is less invasive, but is still costly and has slow refresh speed. Research is still ongoing: there is no modality that has been proven to be effective for localizing microrobots generally.

# 4 Open Problems

## 4.1 Challenges & Open Problems in Robotics for MIS, SPA and NOTES

Minimally invasive surgery in confined spaces has advanced significantly in the last 15 years. Despite the significant progress made, there are still key technical hurdles to successful adoption and deployment of robotic systems in less invasive intraluminal and single port access surgeries. These challenges are hereby summarized primarily based on [6] with suggestions for future areas of research.

The first unmet challenge is still the lack of formal design methodology of robots for MIS, SPA and NOTES. Because engineers and surgeons are still at exploratory stages of these surgical paradigms, designers are often forced to rely on ad-hoc design decisions with trial and error. There is a significant lack of literature documenting anatomical manipulation requirements including forces required for tissue retraction, blunt dissection, and required minimal reachable workspace and dexterous workspace for a given target organ. More works on generating organ-specific tissue models and digital shared sets would greatly facilitate robotics research in the formation of a formal design methodology. The Visible Human Project by the National Library of Medicine is a good start in terms of curating a 3D model of a male and female adult anatomy, but it is still not detailed enough or formed in a model that can be easily annotated with tissue characteristics including stiffness.

The second major challenge preventing the successful deployment of new surgical systems is the fact that new surgical systems for SPA, NOTES and intraluminal surgery generally push the boundaries of the traditional design space. The use of wire-actuated, snake-like and continuum robots has facilitated excellent first stage proof of concept systems, but our current understanding in modeling frictional and actuation losses in these systems is still a hurdle. Simple design questions such as determination of dynamic motion bandwidth by some of these new robot architectures will often result in very hard problem formulations in terms of dynamics, mechanics and control modeling and design. Alternative approaches for use of sensory information for mixed feedback control [42] or the recently proposed model-less control framework [131] are promising new approaches that remain to be further explored. More research in the area of control and modeling of these robots is essential for obtaining high quality motion and force control of these robots.

Besides robot design, there still are fundamental challenges in terms of human-robot interaction, sensing and high-level telemanipulation control. These challenges include the fact that we currently do not have a good framework for designing telemanipulation master devices and user interfaces that are suitable for the highly articulated and branched/multi-arm robot architectures needed to address the requirements of NOTES, SPA and intraluminal surgery. This necessitates the rethinking of how high-level telemanipulation interfaces should be used to help surgeons achieve the surgical outcomes for their patients. Even though there have been works on cooperative manipulation of surgical instruments using semi-active robots such as the Steady-Hand robot[132] or the Acrobot[133], these concepts of human-robot interaction using assistive control laws (known as virtual fixtures[134]) are hard to translate within the context of SPA, NOTES and intraluminal surgery. The difficulty arises from the fact that surgical robots for these new surgical paradigms have to interact with the anatomy along their entire length. As a result, there is a new need for expanding the framework for defining these virtual fixtures to take into account constraints along the robot body. More importantly, there is a need to define new path planning and control and sensing strategies and technologies to allow the in-vivo characterization of the allowable motion space of these robots so that virtual fixtures can be defined in order to assist the user to safeguard the anatomy. Finally, just as in open surgery where multiple surgeons can collaborate of a surgical task, there is a need for new telemanipulation frameworks allowing effective collaboration of at least two surgeons - despite the fact that they have sensory and perception deficiencies regarding the nature of the robot interaction with anatomy in points that are outside the visual field of these robots.

At the level of sensory acquisition and feedback, there have been many works (e.g. [135, 136, 66, 63]) demonstrating the importance and value of force feedback to the users. Some of these works were partly inconclusive due to the quality of the haptic feedback [137], but it is clear that having high quality force feedback is useful and helpful to surgeons to achieve consistent forces of interaction with the tissue and for uniform knot tying, to safeguard against accidental trauma and to help lo- calize tumors. The vast majority of surgical systems today still do not have force feedback. While this issue is not critical in multi-port MIS, it is highly important in NOTES, SPA and intraluminal surgery where the visual perception barriers are even stricter. There have been some recent results in obtaining indirect estimation of forces on continuum and surgical robots [86, 76, 138, 139, 140] or using direct sensing via miniature force sensors[73, 141]. However, these exploratory solutions have not made it into clinical practice either due to cost and sterilization limitations of ded- icated sensory technology or because of complexity of the indirect force estimation algorithms and the uncertainty encumbered in modeling and accounting for friction. In addition to difficulties in manipulation, robotic systems for deep and narrow spaces present difficulties in visualization. These difficulties arise since, in such systems, the endoscope axis and the axis of the surgical access channel lie almost parallel to each other and with a small offset. As a result, the visual field of an endoscope can be easily occluded when the tools move in front of it. During manual MIS, surgeons can use angled lens endoscopes and can rotate the lens to shift the field of view (FOV) to follow the tool tip. This problem is exacerbated when using NOTES/SPA systems because the body of the continuum robot often emanates from the access channel in close proximity to the tip of the endoscope. This creates severe problems of visual occlusion when the continuum robot is telemanipulated. Prior works relevant to camera manipulation FOV management include Reiter and Allen [142] who presented automatic tracking of the arms of the IREP SPA system, but did not consider how to control the continuum arms of the IREP to minimize visual occlusion. Baumann et al. [143] applied a modified probabilistic roadmap method (PRM) to penalize the motions of an articulated serial robot that block the line of sight of an eye-in-hand camera while reaching the visual target. Leonard et al. [144] used PRM and a dynamic collision checking algorithm to plan occlusion-free motions for industrial robots. They modeled the FOV of an eye-to-hand (stationary) camera as a quadtree of frustums and applied adaptive dynamic collision checking algorithms to avoid colliding pre-selected pixels in the view frustum of the camera. Despite these works, the problem of automatic and intelligent strategies for manipulation of endoscopes while minimizing visual occlusion and user disorientation remains an open problem.
Finally, many of the surgical paradigms still fail due to lack of perception and in-vivo sensory information allowing the surgeons to correlate the surgical scene with pre-operative plans. To overcome the fact that organs shift and swell during surgery, recent works such as [145, 146, 147, 148] have started to explore the use of in-vivo model update based on adapting a pre-operative to a model created using sensory data including force, contact location and stiffness. These approaches complement prior works on using geometric scanning and registration of organs (e.g. [149, 150, 151]). Despite progress made in these works, the problem of incorporating in-vivo sensory data to guide and improve the surgical process still stands unsolved.

## 4.2 Challenges & Open Problems in Magnetic Actuation

Despite progress made in magnetically actuated robotics there are several key challenges remaining. The following is a listing of these challenges, primarily based on [87]. While animal trials have been undertaken for some magnetically actuated systems, there are a variety of issues that need improvement before the adoption of these technologies in clinically deployable systems. Magnetic field strength for magnetic coupling depends heavily on the distance between the driving source and the driven magnet, which means high variability depending on abdominal wall thickness and patient obesity levels. High field-strength electromagnetic coils require many wire wrappings and high current, which may require cooling systems that unfavorably increase system size. While significant work in modeling of specific magnetic interactions have been undertaken, further investigation in the flexibility of the anatomical mounting method needs to be done to have a better understanding of these systems' interactions with flexible internal anatomy. Further, there are design problems in placement of multiple coupled magnetic fields in close locations for higher-DOF tasks. Shielded designs and careful design and modeling of magnetic field interaction will need to be undertaken to improve performance. Current methods also would be improved by better feedback sensors to measure position or velocity of internal motion, but size and sterilization requirements makes the integration of such sensors difficult. In their absence, more complex controllers may be required to perform simultaneous estimation during operation.

## 4.3 Challenges & Open Problems in Medical Capsule Robotics

While capsule pills have succeeded in providing visualization capabilities, most commercial systems have done little to move beyond diagnostics to explore capsule therapeutics. Shamshudin et al's review [104] highlights some of the major challenges of the field and room for future growth: Biopsy or other therapeutic capabilities would greatly enhance the abilities of capsule systems. Currently, systems have shown the possibility of therapeutic bacteria or drug release using applied magnetic fields or shape memory alloys, but there is still research ongoing to make these and similar systems realizable clinically. Ex vivo experiments have performed biopsies with shape memory allow or magnetic actuation methods, but again only in a preliminary fashion.

A major restriction of capsule systems is the reliance on batteries for power, which limits the functionality of a given capsule design. Minimizing energy usage or improving energy storage may help, but another alternative is to recharge the capsule using inductive power transfer. Power transfer rates have varied from 150mW to 400mW, which can support imaging and video transmission [104].

While we have summarized many attempted locomotion strategies above, there are none that have undergone full clinical trials in humans. There are still open research questions in the control and localization of capsules in realistic biological environments. The tortuous passages of the intestinal tract pose a unique challenge to control motion while not damaging the internal anatomy or getting stuck in the folds of the flexible intestines.

## 4.4 Challenges & Open Problems in Medical Microrobotics

While there are many promising technologies developed in design and control of microrobots, there are still a wide range of technological hurdles before their adoption in routine clinical practice. For more complete reviews of the state of the art and challenges in microrobotics, see [124, 129, 130].

The key challenges and open problem areas include power, biocompatibility, localization and locomotion. Finding power sources for micro robots is still a major hurdle. For true "micro" robots, generally there is no space for power storage or transmission, so external or anatomical energy sources are more often relied upon. In addition, new work will need to be done on long-term biocompatibility of microrobots moving inside a variety of cavities with different, often corrosive, biological fluids. For certain applications, biodegradable systems may be required for safe use because they will need to either degrade or be removed from the body, which is much more difficult than the insertion process. In contrast with millimeter scale robots where localization is possible with EM tracking or other methods, for non-visualizable microrobots, the main methods of localization rely on expensive imaging modalities such as MRI, CT, or PET scanning. In specific organs, visual tracking can be used, for example in the eye, but this is inherently limited to organs that can be directly visualized. An alternative to tracking individual microrobots is the use of swarms of microrobots. While this adds new dimensions of complexity in their control and coordination, it offers new avenues of their use in a variety of applications. Locmotion of microrobots has been proven in ex-vivo settings for a limited types of fluids and tissues, but locomotion of microrobots against blood flow remains a challenge. Furthermore, the challenges in real-time localization hinder the option of steering these microrobots when moving with blood flow or within other bodily fluids.

Because of the current nascent nature of the technology, current microrobots are limited in proven application scenarios. The difficulties in reliable localization and locomotion keep many of the questions surrounding how these microrobots may be effectively used for diagnostics and therapy unanswered. While the technology is promising, there is still a long road to commercial systems in the microrobotic realm.

# 5 Conclusion

Offering minimally invasive intervention, diagnostics and drug delivery in deep anatomical spaces promises significant benefits to patients, but at the cost of significant technical hurdles to surgeons and engineers attempting at offering solutions leading to this goal. Several technologies including high-dexterity snake-like robots, continuum robots and dedicated systems for single port access surgery and natural orifice surgery have been discussed. With the exception of achieving basic capabilities of manipulation in confined spaces, current systems are subject to many design constraints leading to difficulties in modeling, sensing and control. In addition, the context of performing surgery using these highly dexterous systems in deep surgical sites presents unique challenges to situational awareness of surgeons. These challenges have been identified primarily as challenges due to limited perception of the robot and surgical site and due to indirect routes of access typical of natural orifice surgery and due to the high number of degrees of freedom in such systems. Technologies such as magnetically actuated robots and capsule robots have enabled new approaches to intervention within the patient body. Despite recent progress made there are several unmet challenges in terms of actuation, sensing and localization of these systems within the body. Microrobots offer new opportunities for targeted drug delivery, yet their full potential for other applications remains unlocked due to challenges in power, localization, locomotion and biocompatibility.

# DISCLOSURE STATEMENT

The authors' affiliations were cautiously managed to prevent any influence on opinions expressed in this paper. Nabil Simaan is a co-founder of Continuum Therapeutics - a company aiming at translating natural orifice surgery to clinical practice. He also licensed technologies to Intuitive Surgical, Titan Medical and Auris Medical. He currently holds no stock in any of these companies and has no financial conflict of interest.

# ACKNOWLEDGMENTS

Work reported in this paper has been funded by Vanderbilt University internal funds and in part by NSF#IIS-1327566. The opinions expressed in this paper are those of the authors and do not represent the National Science Foundation.